\documentclass[runningheads]{llncs}
\usepackage{graphicx}
\usepackage[dvipsnames]{xcolor}
\usepackage{subcaption}

\begin{document}
\title{Using Large Language Models for OntoClean-based Ontology Refinement \thanks{This work is a follow-up to the effort done by Team H in "Knowledge Prompting Hackathon" held in August 2023 at King's College London. Full list of team members include (by alphabetical order): Kaveh Aryan, Sarthak Baral, Alba Morales Tirado, Neil Vetter, Youssra Rebboud, Yihang Zhao. Team H was supervised by Prof. Elena Simperl.}}

% \thanks{This work is a follow-up to the effort done by Team H in "Knowledge Prompting Hackathon" held in August 2023 at King's College London. Full list of team members include (by alphabetical order): Kaveh Aryan, Sarthak Baral, Alba Morales Tirado, Neil Vetter, Youssra Rebboud, Yihang Zhao. Team H was supervised by Prof. Elena Simperl.}

% https://docs.google.com/document/d/1L2Q5Szrb4xj8GJQPWbGecCt2hU3JXdA79UeRdgdjf8g/edit#heading=h.45v0nzrpg102

\author{Yihang Zhao\inst{1}\orcidID{0009-0009-2436-8145} \and
Neil Vetter\inst{2}\orcidID{0009-0000-6076-1408} \and
Kaveh Aryan\inst{1}\orcidID{0009-0003-7245-4677}
}

\authorrunning{Y. Zhao et al.}

\institute{King's College London, London, UK \\ \email{\{yihang.zhao, kaveh.aryan\}@kcl.ac.uk} \and
Universität zu Köln, Köln, Germany \\ \email{nvetter3@smail.uni-koeln.de}
}

\maketitle              % typeset the header of the contribution

\begin{abstract}
This paper explores the integration of Large Language Models (LLMs) such as GPT-3.5 and GPT-4 into the ontology refinement process, specifically focusing on the OntoClean methodology. OntoClean, critical for assessing the metaphysical quality of ontologies, involves a two-step process of assigning meta-properties to classes and verifying a set of constraints. Manually conducting the first step proves difficult in practice, due to the need for philosophical expertise and lack of consensus among ontologists. By employing LLMs with two prompting strategies, the study demonstrates that high accuracy in the labelling process can be achieved. The findings suggest the potential for LLMs to enhance ontology refinement, proposing the development of plugin software for ontology tools to facilitate this integration.

\end{abstract}
\section{Introduction} \label{section:introduction}
In semantic web technologies, the construction of ontologies is a foundational first step. However, the creation of an ontology is a first step, and a critical and often challenging step is the evaluation and refinement of the constructed ontology. This paper focuses on the OntoClean methodology, which is renowned for addressing the metaphysical quality of ontologies through the analysis of their hierarchical structures.

OntoClean is implemented through a two-step process: the primary step involves assigning a predetermined set of OntoClean metaproperties to the classes within the ontology. The subsequent step, which is more straightforward, entails verifying whether the OntoClean constraints, predicated on these assigned labels, are satisfied. The initial labelling phase represents the principal challenge in applying OntoClean due to its demand for philosophical expertise. Furthermore, reaching a consensus on the labelling of specific classes can prove difficult for ontologists.

Given the proficiency of large language models (LLMs) in a broad range of text comprehension and reasoning tasks, we suggest utilising them to streamline the labelling step. To explore this potential, we employed two prompting strategies, namely a zero-shot "bare" prompt and a prompt that included meta-properties' documentation (in-context learning prompt). Using these prompt, we experimented with GPT-3.5 and GPT-4 LLMs. We evaluated different strategies using two carefully curated benchmarks. Our best results, achieved on GPT-4, show a very high accuracy across the range of OntoClean meta-properties. This illuminates the feasibility and efficiency of integrating LLMs into the ontology refinement process, by for instance, development of plugin software for common ontology software.

The rest of the paper is structured as follows: Section \ref{section:related-work} surveys the previous work on using LLMs in ontology refinement and also reviews OntoClean methodology. Section \ref{section:methods} introduces the proposed pipeline, benchmarks, and chosen prompting strategies. Section \ref{section:results} presents the results, and, finally, Section \ref{section:conclusion} concludes the paper.

\section{Related works} \label{section:related-work}

\subsection{Ontology refinement} \label{section:refinement}
Ontology (and knowledge graph) refinement is a critical process undertaken after the initial construction phase --a phase that might be conducted by an organisation different from the one that developed the ontology \cite{cimiano_knowledge_2017}. This step involves taking an existing ontology as input and addressing various issues it may have, especially, incompleteness and inaccuracies (errors). Refinement aims to enhance the ontology by adding missing entities or relationships (completion) and ensuring that existing relationships accurately reflect the real-world (correction). Refinement can occur at different levels: the conceptual level (T-box) concerns the taxonomy and structure of the ontology, and the instance level (A-box) deals with individual instances. In our study, we concentrate on OntoClean (discussed in \ref{section:ontoclean}) which is a T-box level refinement method. Additionally, there are other methods at the T-box level: \cite{beydoun_how_2011} provides formal foundations for assessing the quality of a taxonomy within an ontology and suggests a semi-automatic methodology for developing maintainable taxonomies. \cite{corcho_pattern-based_nodate} identifies inconsistencies within an ontology and \cite{masuda_ontology_nodate} focuses on a specific type of relations, namely, 'is-a' relations.

\subsection{Large language models and prompting techniques} \label{section:llm-prompting}
A large language model is a statistical model that captures the syntax and semantics of one or more target languages. The latest advancements use deep neural networks, mainly the transformer architecture \cite{vaswani_attention_2023}. A full transformer is a sequence-to-sequence model, with an encoder to process the input and a decoder to generate output. Milestones in the transformer architecture are BERT, which pushed the state of the art on several natural language processing tasks and benchmarks \cite{devlin_bert_2018}, and GPT, Bard, and Cluade which, as powerful chatbots, gave the language models their publicity \cite{brown_language_2020}\cite{borji_battle_2023}. Interaction with a chatpot is done by providing the model a task description, or "prompt," to generate an appropriate response. Various prompting strategies exist. In this paper we used zero-shot prompting where a bare description of the task is provided, and a variant of in-context learning where we provide a demonstrative definitions in the prompt to steer the model to achieve better performance.

\subsection{Large language models for ontology refinement} \label{section:refinement}
Recent advancements in ontology refinement have featured the use of large language models, especially BERT and its derivatives. \cite{zeng_enhancing_2021} employs SciBERT in GenTaxo for taxonomy completion by identifying areas requiring new concepts and generating their names, focusing on contextual embeddings from relational data rather than traditional corpus-based embeddings. Similarly, \cite{chen_contextual_2023} presents BERTSubs, a method for automating OWL ontology construction and curation using BERT. This method generates contextual embeddings of ontology classes, integrating class context and logical existential restrictions through custom templates for accurate prediction of various subsumers, including named classes and existential restrictions from the same or different ontologies. \cite{liu_concept_2020} focuses on automating the prediction of IS-A relationships in ontologies using BERT's Next Sentence Prediction capability. Their approach involves converting the neighborhood network of a concept into “sentences” and refining training data with ontology summarization techniques. \cite{shi_subsumption_2023} explores the characteristics of e-commerce taxonomies and introduces a new subsumption prediction method for this sector, utilizing the pre-trained language model BERT and integrating textual and structural elements of a taxonomy.

\subsection{OntoClean} \label{section:ontoclean}
In the domain of ontology evaluation and refinement, the OntoClean methodology, pioneered by Guarino and Welty \cite{guarino_overview_2009}, stands out as a principled framework for assessing ontology quality in terms of completeness and accuracy. It leverages philosophical insights to establish meta-properties that classes within an ontology should exhibit, thus facilitating a rigorous analysis of the ontology's structure.
\subsubsection{Meta-properties} \label{section:ontoclean-metaproperties}
The core meta-properties of OntoClean are Identity (I), Unity (U), Rigidity (R), and Dependence (D). They serve as the foundation for the analysis, each contributing to a nuanced understanding of class characteristics and their implications for ontology design.
\begin{itemize}
    \item \textbf{Identity (I)} addresses the criteria that determine the distinctiveness of instances within a class, ensuring that entities can be uniquely identified (+I) or acknowledging the absence of such criteria (--I). This principle is vital for distinguishing when instances are identical or disparate, thereby influencing the ontological representation of concepts like "Person" (marked +I for unique identities) versus "Water" (marked -I due to indistinguishability).
    \item \textbf{Unity (U)} examines whether instances of a class are perceived as wholes. It differentiates between indivisible entities (+U) and those that are collections of distinct parts (--U). Additionally, anti-unity ($\sim U$) identifies classes where instances inherently lack cohesion, further distinguishing between entities that cannot be perceived as wholes.
    \item \textbf{Rigidity (R)} pertains to the necessity of class membership across an instance's existence, categorising classes as rigid (+R) if membership is constant, or non-rigid (--R) if it can change over time. Within non-rigid, anti-rigid ($\sim$ R) properties further specify classes where membership is non-essential and alterable without affecting the instance's existence or identity.
    \item \textbf{Dependence (D)} investigates the existential reliance of instances on entities from other classes, marking classes as dependent (+D) when such a relationship exists, or independent (--D) when it does not.
\end{itemize}

\subsubsection{Constraints} \label{section:ontoclean-constraints}
We use constraints on the meta-properties to maintain the logical consistency and integrity of ontological taxonomies. These are the key constraints for label combinations and inheritance:
\begin{itemize}
    \item Identity Constraints: If a class carries an identity criterion (+I), subclasses under it must also carry this identity criterion (+I).
    \item Rigidity Constraints: If a class is marked anti-rigid ($\sim R$), then its subclasses must also be marked anti-rigid ($\sim R$).
    \item Unity Constraints: When a class is defined with a unity criterion (+U), its subclasses must adhere to the same unity criterion (+U).
    \item Anti-Unity Constraints: If a class is characterized by anti-unity ($\sim U$), this trait extends to its subclasses.
    \item Dependency Constrains: If a class is deemed dependent (+D) on some external class, then its subclasses must also be externally dependent (+D).
\end{itemize}
In addition to these constraints we adhere to the following assumptions on identity, highlighting essential aspects of entity identification within an ontology:
\begin{itemize}
    \item Sortal Individuation states that every entity must carry at least one property with an identity criterion (+I), aligning with the principle “No entity without identity”, ensuring every entity is uniquely identifiable.
    \item Sortal Expandability suggests that if an entity is categorized under multiple properties at different times or contexts, it must also belong to a more general property that provides a unifying identity criterion (+I), making sure the entity's identity is consistently identifiable across various classifications.
\end{itemize}
These rules ensure that ontological structures maintain logical coherence.

\iffalse
\subsubsection{Limitations} \label{section:ontoclean-limitations}
While the OntoClean methodology offers a structured and principled approach to ontology evaluation, it is not without its constraints. One of the primary limitations is the significant reliance on manual analysis and the need for deep philosophical understanding, as well as domain-specific knowledge. This requirement can make the application of OntoClean resource-intensive and potentially inaccessible to non-experts in ontology design or those lacking in philosophical training. Additionally, the methodology's emphasis on meta-properties and their philosophical underpinnings may not directly address all practical concerns of ontology use, such as operational efficiency, scalability, or integration with existing data systems. Furthermore, OntoClean focuses primarily on the taxonomic structure and the conceptual clarity of an ontology, possibly overlooking aspects related to dynamic properties of instances, such as their behaviour or interactions over time. These constraints highlight the necessity for complementary approaches and tools that can mitigate these challenges, providing a more holistic framework for ontology development and evaluation in complex and varied application domains.
\fi

\section{Methods} \label{section:methods}
\subsection{Process} \label{section:methods-prompts}
The input for the pipeline consists of the unlabeled classes of the ontology, along with the underlying class hierarchy. The LLM is employed to assign labels to each class, based on the specific context that the class occupies within the ontology. To achieve this, the model must reason for each class, taking into account both the inherent properties of the class and its relationships with other classes. This includes considering the dependencies between the class in question and the classes upon which it relies, as well as those that rely on it. 
\begin{figure}
\includegraphics[scale=0.31]{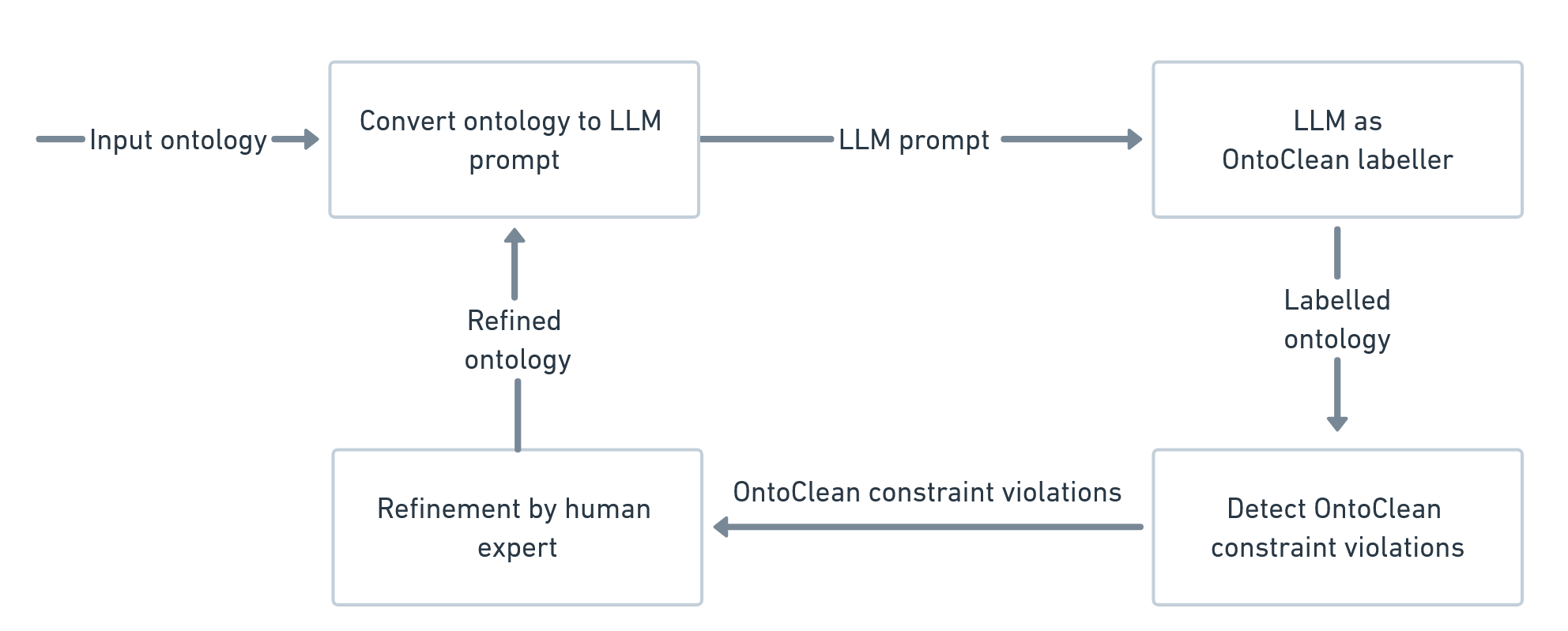}
\caption{Pipeline for the Human/AI OntoClean refinement process.} \label{fig:pipeline}
\end{figure}
We can now identify instances in the labelled ontology that violate the constraints we set for the labels. In the next step the researcher must assess how accurate the labelling is. This is a 'human-in-the-loop' approach, where the researcher's feedback is crucial. If the labelling shows low accuracy, the researcher has two options: they can either tweak the initial prompt or provide the LLM with additional guidance and resubmit the partially mislabelled ontology.

\textbf{Metrics} We assess the quality of the class labels by looking at each meta-property (I, U, R, D) individually, calculating an accuracy score. This allows us to evaluate each meta-property separately. If we find that one property is significantly weaker (or stronger) than the others, we can use this insight to adjust our approach. We do this by either expanding on the description of that property if it's weak or by using a similar description for the other properties, if it's strong.
\subsection{Two Benchmarks} \label{section:methods-benchmarks}
\subsubsection{The Mini Pizza Ontology}
The Mini Pizza Ontology is a simplified ontology designed primarily for educational purposes, focusing on the domain of pizzas and their ingredients. It serves as an exemplary model to demonstrate the principles of ontology development, including class hierarchy, properties, and instances. It covers various types of pizzas, ingredients (like toppings and bases), and the relationships between them. The Mini Pizza Ontology is part of the larger Pizza Ontology developed at the University of Manchester for the TONES Ontology Repository Project.

\subsubsection{The Upper Ontology}
Upper Ontologies (also called top-level ontologies) are ontologies designed to provide a high-level framework that can be used to support a wide range of domain-specific ontologies. They typically include very general concepts such as object, event, or relation, which are applicable across various domains. They provide a basic structure that helps different specialized ontologies work together, making it easier to share data and knowledge between various systems and fields.

\subsection{Prompting} \label{section:methods-prompts}
\begin{figure}[h]
\includegraphics[width=\textwidth]{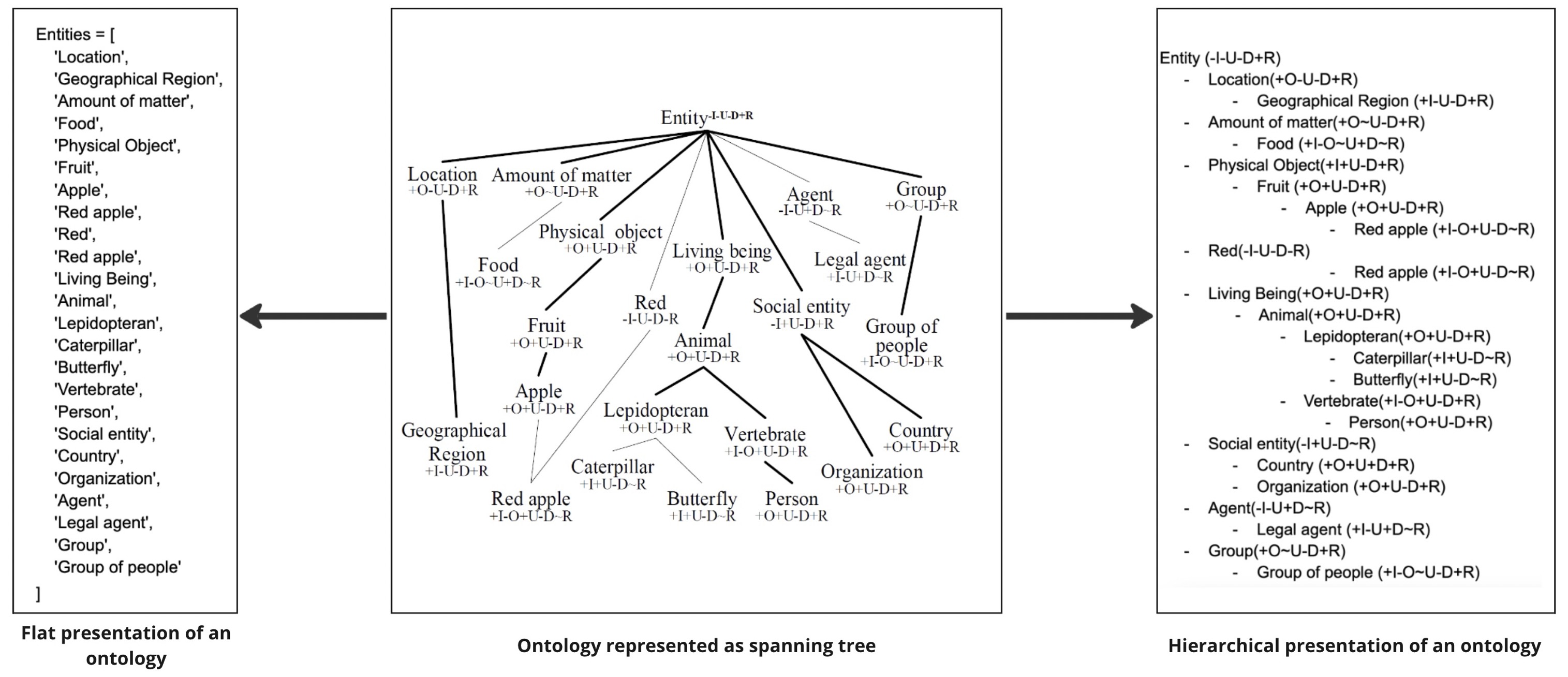}
\caption{Flat and hierarchical representations of an example ontology.} \label{fig:representation}
\end{figure}

For the task of labelling an ontology with OntoClean meta-properties, we experimented with two prompts: a bare prompt, and an in-context learning prompt, where the LLM was reminded of the definitions of OntoClean meta-properties. See Table \ref{table:prompts} for details. This prompt was concatenated with a textual representation of the ontology, which was either flat or hierarchical. Flat presentation is a list of entities present in the ontology. For hierarchical presentation, we first chose a random spanning tree of the ontology graph and then presented as a tab-indented text. See Fig. \ref{fig:representation} for an example.

\begin{table}
\caption{Prompts for labelling ontologies with OntoClean meta-properties} \label{table:prompts}
\begin{tabular}{|p{0.15\textwidth}|p{0.84\textwidth}|}
\hline
\textbf{Prompt type} & \textbf{Prompt text} \\
\hline
"Bare" (Zero-shot) & Label this ontology with OntoClean criteria. \\
\hline
In-context learning & Label this ontology with OntoClean criteria. Rigidity='Rigidity is based on the notion of essence. A concept is essential for an instance iff it is necessarily an instance of this concept, in all worlds and at all times. ...' \\
\hline
\end{tabular}
\end{table}

\section{Results} \label{section:results}
\begin{figure}[h]
    \centering
    \begin{subfigure}[b]{0.35\textwidth}
        \includegraphics[width=\textwidth]{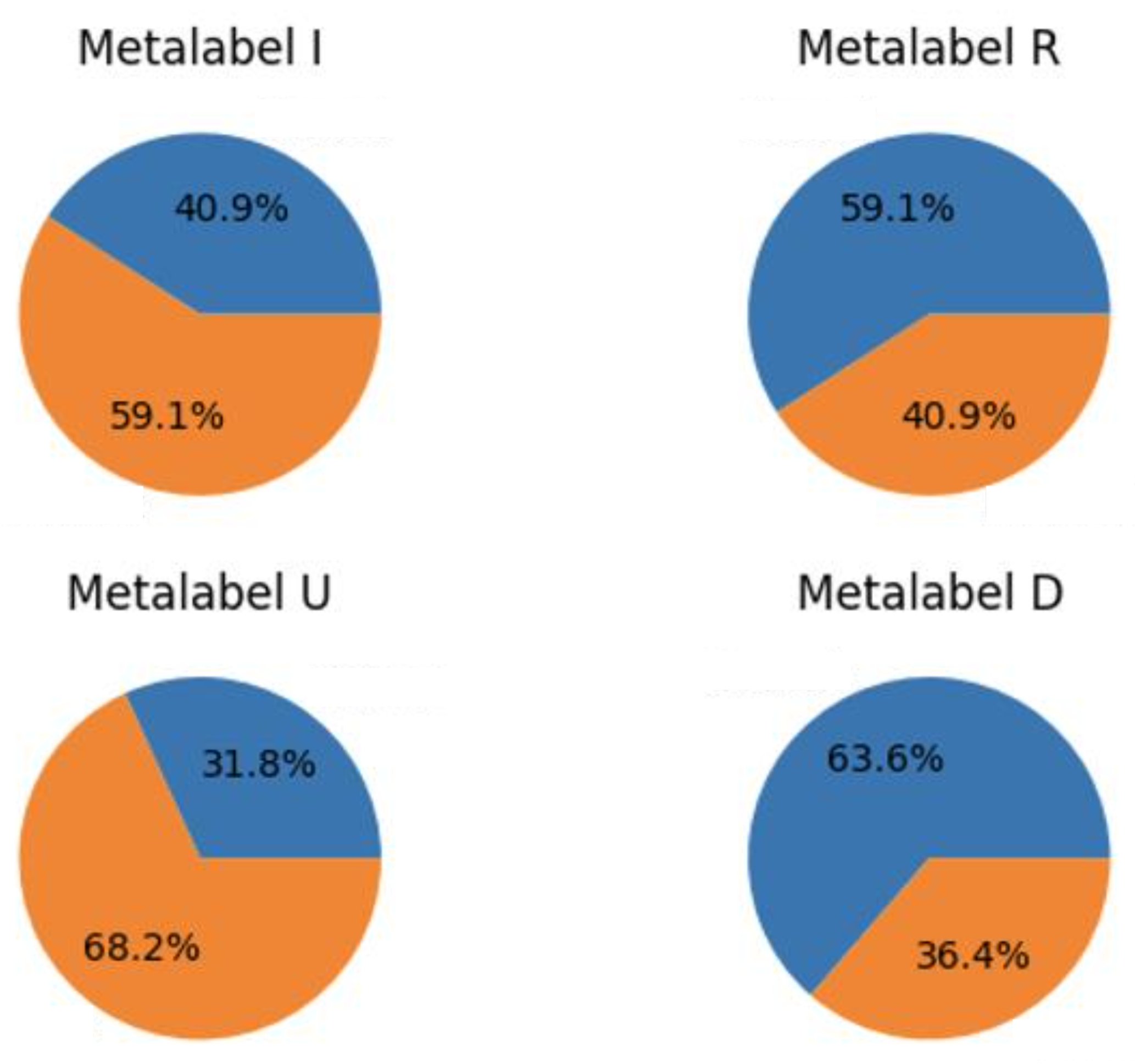}
        \caption{Zero-shot (GPT-3.5)}
        \label{fig:gpt4a}
    \end{subfigure}
    \hfill % adds horizontal space between the figures
    \begin{subfigure}[b]{0.35\textwidth}
        \includegraphics[width=\textwidth]{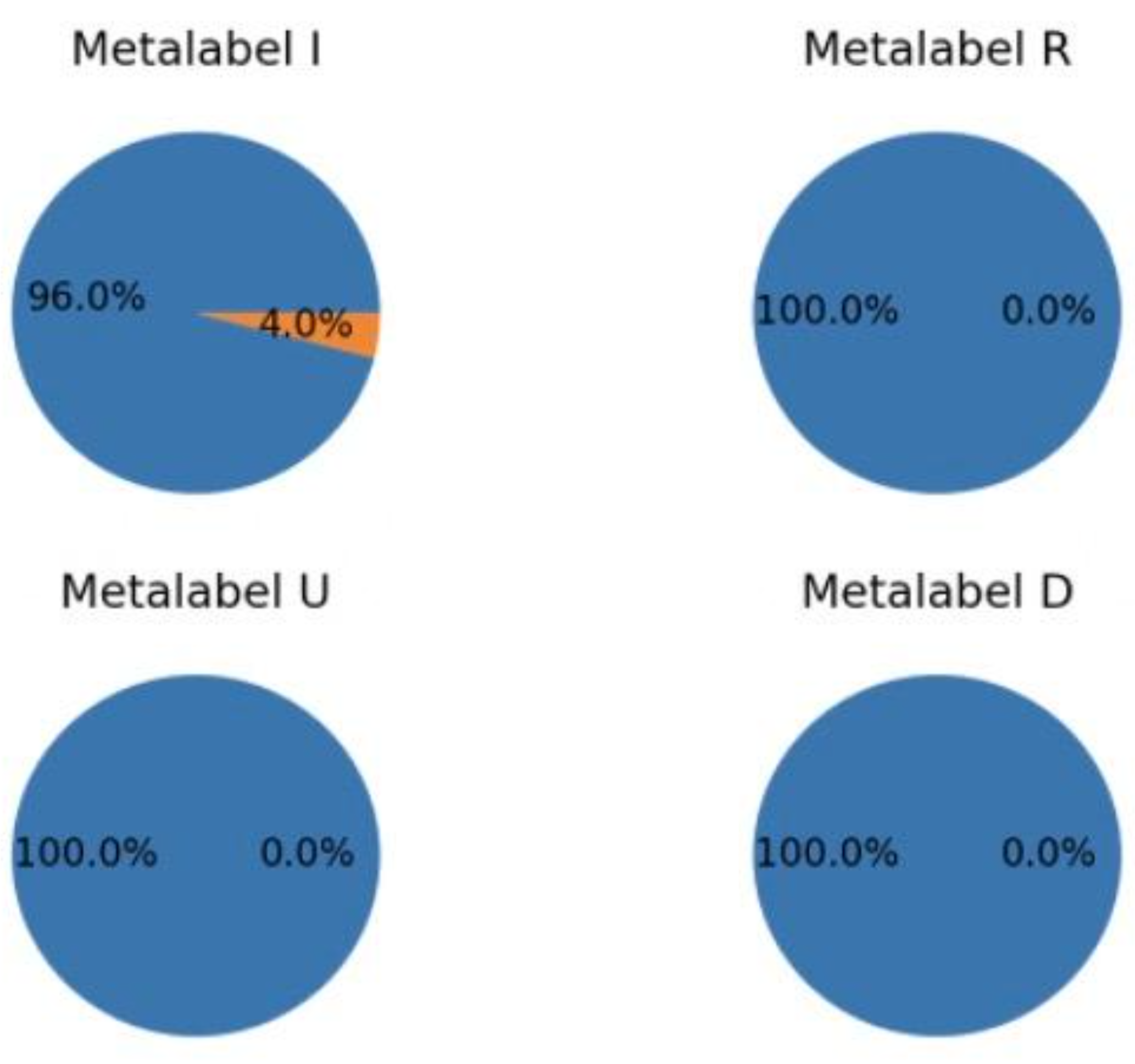}
        \caption{Zero-shot (GPT-4)}
        \label{fig:gpt4b}
    \end{subfigure}
    \par\bigskip % adds vertical space between the rows of figures
    \begin{subfigure}[b]{0.35\textwidth}
        \includegraphics[width=\textwidth]{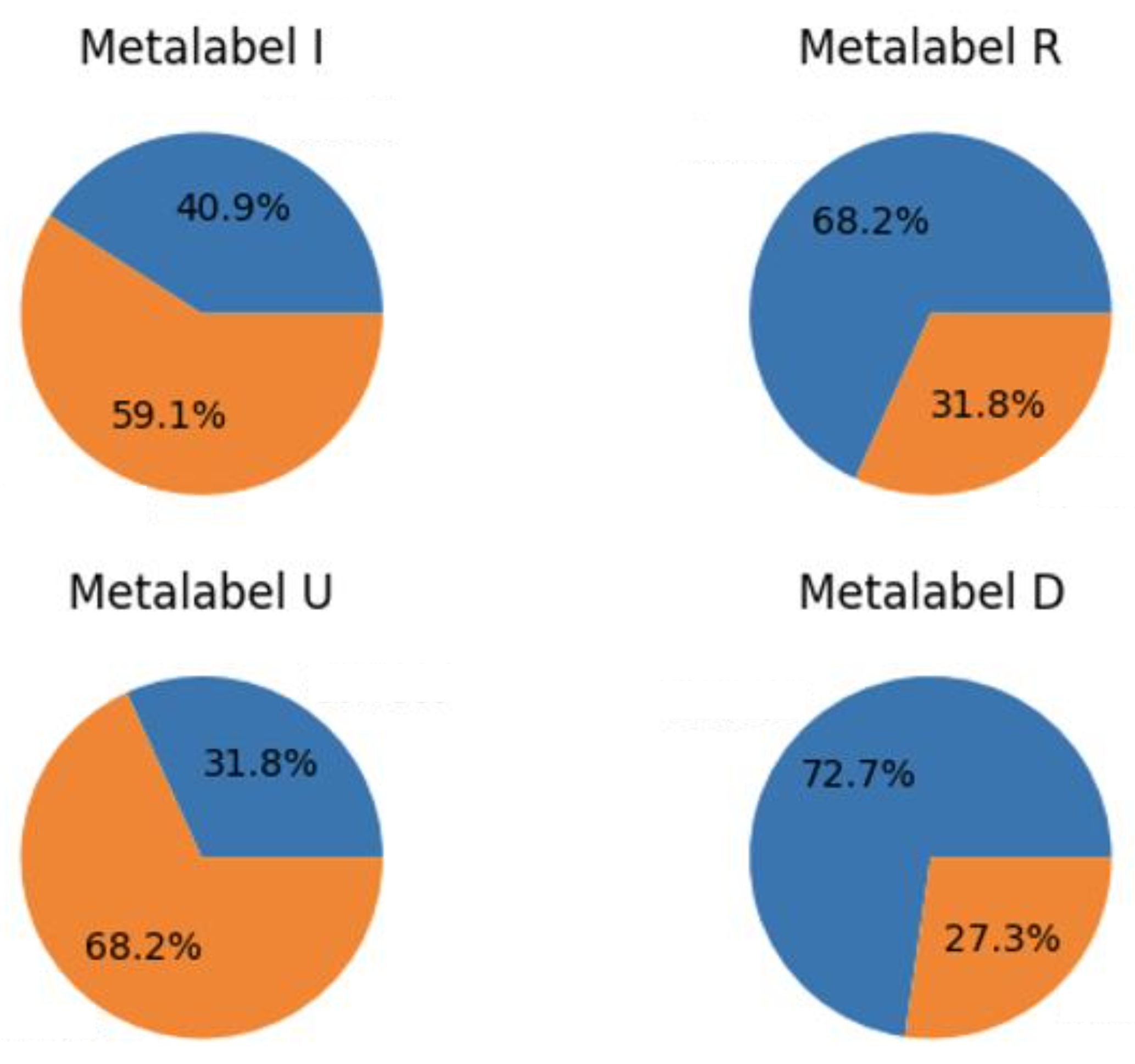}
        \caption{In-context learning (GPT-3.5)}
        \label{fig:gpt4c}
    \end{subfigure}
    \hfill
    \begin{subfigure}[b]{0.35\textwidth}
        \includegraphics[width=\textwidth]{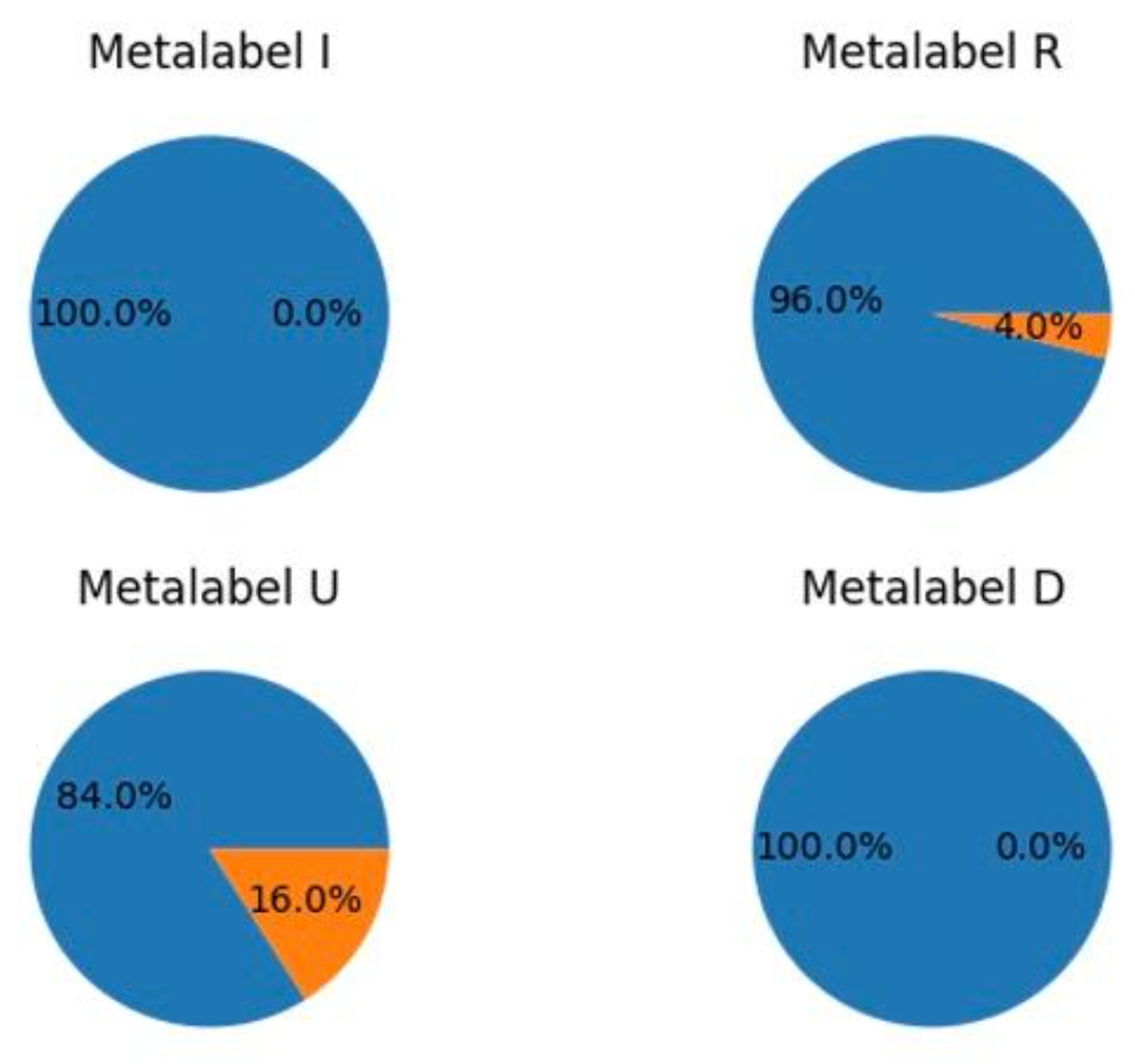}
        \caption{In-context learning (GPT-4)}
        \label{fig:gpt4d}
    \end{subfigure}
    \caption{Experimental results for different prompts and LLMs. Blue indicates correct and orange incorrect labels.}
    \label{fig:Experiment}
\end{figure}

In our research, we evaluated both the Upper Ontology and Mini Pizza Ontology using the GPT-3.5 and GPT-4 models, incorporating zero-shot and in-context learning approaches with both flat and hierarchical representations of the ontologies as input. We executed 30 trials for each permutation of ontology, model, prompting approach, and representation style. Following this, we combined and compared the results from both ontologies to analyze the accuracy rates between GPT-3.5 and GPT-4 across the different prompting techniques and representations styles. Our findings indicated that using flat representations as inputs for the ontologies did not yield satisfactory results with any of the language models or prompting techniques; the outputs lacked dependency or hierarchy, rendering them unusable. Consequently, we excluded the results from flat presentations, focusing solely on the more relevant and informative results obtained with hierarchical presentation inputs, as depicted in the subsequent Fig.\ref{fig:Experiment}.

GPT-3.5 showed suboptimal performance, especially in meta properties Identity (I) and Unity (U), with inaccuracy rates ranging from approximately 60\% to 70\%, regardless of the learning context. This consistent underperformance could be attributed to GPT-3.5's limitations in handling abstract ontological concepts and distinguishing unique identifiers or cohesive wholes, observable in both zero-shot and in-context learning scenarios. Nevertheless, a marginal improvement was noted in meta properties Rigidity (R) and Dependence (D). Without in-context learning, inaccuracy rates for these labels hovered around 35\% to 40\%, which decreased to 25\% to 30\% with the application of in-context learning, possibly due to GPT-3.5's relative adeptness in comprehending more concrete and stable class memberships, as well as existential interdependencies.

In contrast, GPT-4 exhibited superior performance across all testing conditions. It notably achieved very low inaccuracy rates of around 4\% in meta properties Identity (I) and Rigidity (R), reflecting its advanced understanding and application of criteria for distinctiveness and class membership stability. However, a notable observation was that during in-context learning, GPT-4 did not display improved performance compared to zero-shot learning in meta properties Unity (U). This could be due to the limited scope and abstract nature of the definitions in ontology domain, which date back to 2004 and lack of clear examples for each label, potentially leading to confusion in the model's interpretation. The abstractness and dated references in OntoClean might have posed challenges for GPT-4 in contextualizing and accurately applying the concepts of Unity.

\section{Conclusion} \label{section:conclusion}
In this paper, we investigated the potential of using LLMs for ontology refinement via the OntoClean methodology. We experimented with various prompts and LLMs. Our results indicate that our most capable model, GPT-4, achieves very high accuracy across the range of OntoClean metaproperties. Based on these promising results, we propose a complete OntoClean-based human-in-the-loop approach. Specifically, we suggest developing a plugin for ontology software, such as Protege, to support OntoClean-LLM integration as outlined in this paper.

\bibliography{teamh.bib}

\end{document}